\documentclass[letterpaper, 10 pt, conference]{ieeeconf}  

\usepackage{times}  
\usepackage{helvet}  
\usepackage{courier}  
\usepackage{url}  
\usepackage{graphicx}  

\usepackage{booktabs} 
\usepackage{bm}
\usepackage[ruled,lined,linesnumbered]{algorithm2e}
\usepackage{algorithmic}
\usepackage{array}
\usepackage{multirow}
\usepackage{enumerate}
\usepackage{subfigure}
\usepackage{dsfont}

\usepackage[hidelinks]{hyperref}
\usepackage{amsfonts}
\usepackage{amsmath}
\usepackage{adjustbox}
\usepackage{blindtext}
\usepackage{fixltx2e}

\graphicspath{ {figures/} }

\IEEEoverridecommandlockouts                              

\title{\LARGE \bf
SqueezeSegV2: Improved Model Structure and Unsupervised Domain Adaptation for Road-Object Segmentation from a LiDAR Point Cloud
}

\author{Bichen Wu$^*$, Xuanyu Zhou$^*$, Sicheng Zhao$^*$, Xiangyu Yue, Kurt Keutzer \\
UC Berkeley \\
\{bichen, xuanyu\_zhou, schzhao, xyyue, keutzer\}@berkeley.edu
\thanks{* Authors contributed equally.}
}

\begin{document}

\maketitle
\thispagestyle{empty}
\pagestyle{empty}

\begin{abstract}
 Earlier work demonstrates the promise of deep-learning-based approaches for point cloud segmentation; however, these approaches need to be improved to be practically useful. To this end, we introduce a new model SqueezeSegV2 that is more robust to dropout noise in LiDAR point clouds. With improved model structure, training loss, batch normalization and additional input channel, SqueezeSegV2 achieves significant accuracy improvement when trained on real data. Training models for point cloud segmentation requires large amounts of labeled  point-cloud data, which is expensive to obtain. To sidestep the cost of collection and annotation, simulators such as GTA-V can be used to create unlimited amounts of labeled, synthetic data. However, due to domain shift, models trained on synthetic data often do not generalize well to the real world. We address this problem with a domain-adaptation training pipeline consisting of three major  components:  1) learned intensity rendering, 2) geodesic  correlation alignment, and 3) progressive domain calibration. When trained on real data, our new model exhibits segmentation accuracy improvements of 6.0-8.6\% over the original SqueezeSeg. When training our new model on synthetic data using the proposed domain adaptation pipeline, we nearly double test accuracy on real-world data, from 29.0\% to 57.4\%. Our source code and synthetic dataset will  be  open-sourced.
\end{abstract}
\section{INTRODUCTION}
\label{sec:Introduction}
Accurate, real-time, and robust perception of the environment is an indispensable component in autonomous driving systems. For perception in high-end autonomous vehicles, LiDAR (Light Detection And Ranging) sensors play an important role. LiDAR sensors can directly provide distance measurements, and their resolution and field of view exceed those of radar and ultrasonic sensors~\cite{moosmann2009segmentation}. LiDAR sensors are robust under almost all lighting conditions: day or night, with or without glare and shadows~\cite{wu2017squeezeseg}. As such, LiDAR-based perception has attracted significant research attention.

\begin{figure}[!t]
\begin{center}
\centering \includegraphics[width=0.9\linewidth]{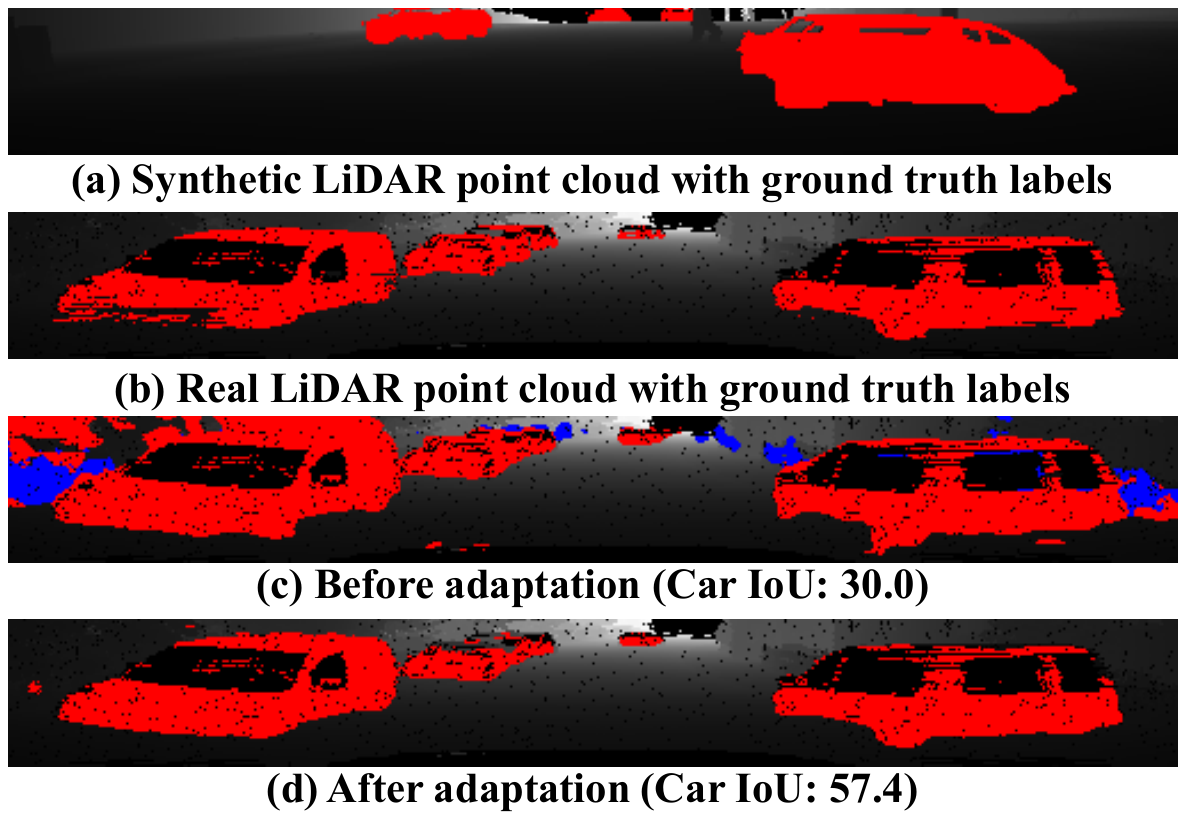}
\caption{An example of \emph{domain shift}. The point clouds are projected onto a spherical surface for visualization (car in red, pedestrian in blue). Our domain adaptation pipeline improves the segmentation from (c) to (d) while trained on synthetic data.}

\vspace{-0.9cm}
\label{fig:DomainShift}
\end{center}
\end{figure}

Recently, deep learning has been shown to be very effective for LiDAR perception tasks. Specifically, Wu et al. proposed SqueezeSeg \cite{wu2017squeezeseg}, which focuses on the problem of point-cloud segmentation. SqueezeSeg projects a 3D LiDAR point cloud onto a spherical surface, and uses a 2D CNN to predict point-wise labels for the point cloud. SqueezeSeg is extremely efficient -- the fastest version achieves an inference speed of over 100 frames per second. However, SqueezeSeg still has several limitations: first, its accuracy still needs to be improved to be practically useful. One important reason for accuracy degradation is \textit{dropout noise} -- missing points from the sensed point cloud caused by limited sensing range, mirror diffusion of the sensing laser, or jitter in incident angles. Such dropout noise can corrupt the output of SqueezeSeg's early layers, which reduces accuracy. Second, training deep learning models such as SqueezeSeg requires tens of thousands of labeled point clouds; however, collecting and annotating this data is even more time consuming and expensive than collecting comparable data from cameras. GTA-V is used to synthesize LiDAR point cloud as an extra source of training data~\cite{wu2017squeezeseg}; however, this approach suffers from the domain shift problem \cite{torralba2011unbiased} -  models trained on synthetic data usually fail catastrophically on the real data, as shown in Fig.~\ref{fig:DomainShift}. Domain shift comes from different sources, but the absence of dropout noise and intensity signals in GTA-V are two important factors. Simulating realistic dropout noise and intensity is very difficult, as it requires sophisticated modeling of both the LiDAR device and the environment, both of which contain a lot of non-deterministic factors. As such, the LiDAR point clouds generated by GTA-V do not contain dropout noise and intensity signals. The comparison of simulated data and real data is shown in Fig.~\ref{fig:DomainShift} (a), (b).

In this paper, we focus on addressing the challenges above. First, to improve the accuracy, we mitigate the impact of dropout noise by proposing the Context Aggregation Module (CAM), a novel CNN module that aggregates contextual information from a larger receptive field and improves the robustness of the network to dropout noise. Adding CAM to the early layers of SqueezeSegV2 not only significantly improves its performance when trained on real data, but also effectively reduces the domain gap, boosting the network's real-world test accuracy when trained on synthetic data. In addition to CAM, we adopt several improvements to SqueezeSeg, including using focal loss \cite{lin2018focal}, batch normalization \cite{ioffe2015batch}, and LiDAR mask as an input channel. These improvements together boosted the accuracy of SqueezeSegV2 by 6.0\% - 8.6\% in all categories on the converted KITTI dataset \cite{wu2017squeezeseg}. 

Second, to better utilize synthetic data for training the model, we propose a domain adaptation training pipeline that contains the following steps: first, before training, we render intensity channels in synthetic data through \textit{learned intensity rendering}. We train a neural network that takes the point coordinates as input, and predicts intensity values. This rendering network can be trained in a "self-supervised" fashion on unlabeled real data. After training the network, we feed the synthetic data into the network and render the intensity channel, which is absent from the original simulation. Second, we use the synthetic data augmented with rendered intensity to train the network. Meanwhile, we follow \cite{morerio2018minimal} and use \textit{geodesic correlation alignment} to align the batch statistics between real data and synthetic data. 3) After training, we propose \textit{progressive domain calibration} to further reduce the gap between the target domain and the trained network. Experiments show that the above domain-adaptation training pipeline significantly improves the accuracy of the model trained with synthetic data from 29.0\% to 57.4\% on the real world test data. 

The contributions of this paper are threefold: 1) We improve the model structure of SqueezeSeg with CAM to increase its robustness to dropout noise, which leads to significant accuracy improvements of 6.0\% to 8.6\% for different categories. We name the new model SqueezeSegV2. 2) We propose a domain-adaptation training pipeline that significantly reduces the distribution gap between synthetic data and real data. Model trained on synthetic data achieves 28.4\% accuracy improvement on the real test data. 3) We create a large-scale 3D LiDAR point cloud dataset, GTA-LiDAR, which consists of 100,000 samples of synthetic point cloud augmented with rendered intensity. The source code and dataset will be open-sourced. 

\section{RELATED WORK}
\label{sec:RelatedWork}
\textbf{3D LiDAR Point Cloud Segmentation} aims to recognize objects from point clouds by predicting point-wise labels. Non-deep-learning methods~\cite{moosmann2009segmentation, douillard2011segmentation, zermas2017fast} usually involve several stages such as ground removal, clustering, and classification. SqueezeSeg~\cite{wu2017squeezeseg} is one early work that applies deep learning to this problem. Piewak et al.~\cite{piewak2018boosting} adopted a similar problem formulation and pipeline to SqueezeSeg and proposed a new network architecture called LiLaNet. They created a dataset by utilizing image-based semantic segmentation to generate labels for the LiDAR point cloud. However, the dataset was not released, so we were not able to conduct a direct comparison to their work. Another category of methods is based on PointNet~\cite{qi2017pointnet, qi2017pointnet++}, which treats a point cloud as an unordered set of 3D points. This is effective with 3D perception problems such as classification and segmentation. Limited by its computational complexity; however, PointNet is mainly used to process indoor scenes where the number of points is limited. Frustum-PointNet~\cite{qi2017frustum} is proposed for out-door object detection, but it relies on image object detection to first locate object clusters and feeds the cluster, instead of the whole point cloud, to the PointNet.

\textbf{Unsupervised Domain Adaptation (UDA)} aims to adapt the models from one labeled source domain to another unlabeled target domain. Recent UDA methods have focused on transferring deep neural network representations~\cite{patel2015visual,csurka2017domain}. Typically, deep UDA methods employ a conjoined architecture with two streams to represent the models for the source and target domains, respectively. In addition to the task related loss computed from the labeled source data, deep UDA models are usually trained jointly with another loss, such as a discrepancy loss~\cite{long2015learning,sun2017correlation,zhuo2017deep,zhang2017curriculum,morerio2018minimal}, adversarial loss~\cite{liu2016coupled,ganin2016domain,tzeng2017adversarial,shrivastava2017learning,bousmalis2017unsupervised,hoffman2018cycada}, label distribution loss~\cite{zhang2017curriculum} or reconstruction loss~\cite{ghifary2015domain,ghifary2016deep}.

The most relevant work is the exploration of synthetic data~\cite{shrivastava2017learning,zhang2017curriculum,hoffman2018cycada}. By enforcing a self-regularization loss, Shrivastava et al.~\cite{shrivastava2017learning} proposed SimGAN to improve the realism of synthetic data using unlabeled real data. Another category of relevant work employs a discrepancy loss~\cite{long2015learning,sun2017correlation,zhuo2017deep,morerio2018minimal}, which explicitly measures the discrepancy between the source and target domains on corresponding activation layers of the two network streams. Instead of working on 2D images, we try to adapt synthetic 3D LiDAR point clouds by a novel adaptation pipeline.

\textbf{Simulation} has recently been used for creating large-scale ground truth data for training purposes. Richter et al.~\cite{playingfordata} provided a method to extract semantic segmentation for the synthesized in-game images. In \cite{drivinginthematrix}, the same game engine is used to extract ground truth 2D bounding boxes for objects in the image. Yue et al.~\cite{yue2018lidar} proposed a framework to generate synthetic LiDAR point clouds. Richter et al.~\cite{playingforbenchmarks} and Kr\"ahenb\"uhl~\cite{philip2018free} extracted more types of information from video games. 

\begin{figure*}[!t]
\begin{center}
\centering \includegraphics[width=0.65\linewidth]{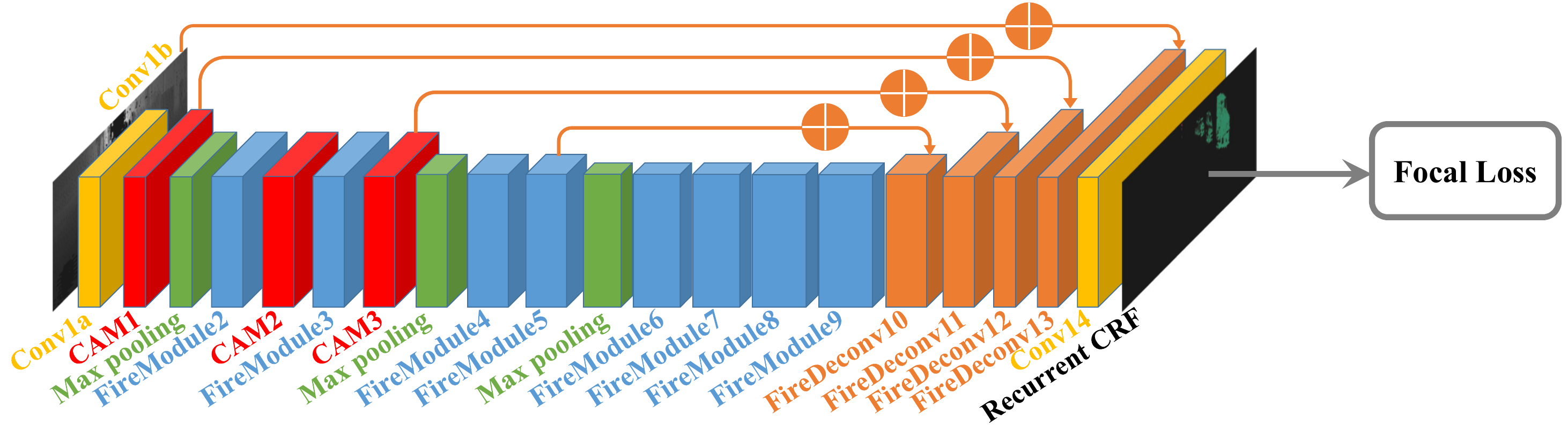}
\caption{Network structure of the proposed SqueezeSegV2 model for road-object segmentation from 3D LiDAR point clouds.}
\label{fig:SqueezeSegV2Framework}
\end{center}
\end{figure*}

\section{Improving the model structure}
\label{sec:Better}

We propose SqueezeSegV2, by improving upon the base SqueezeSeg model, adding Context Aggregation Module (CAM), adding LiDAR mask as an input channel, using batch normalization \cite{ioffe2015batch}, and employing the focal loss \cite{lin2018focal}. The network structure of SqueezeSegV2 is shown in Fig.~\ref{fig:SqueezeSegV2Framework}.

\subsection{Context Aggregation Module}
\label{ssec:CAM}
\begin{figure}[!t]
\begin{center}
\centering \includegraphics[width=0.8\linewidth]{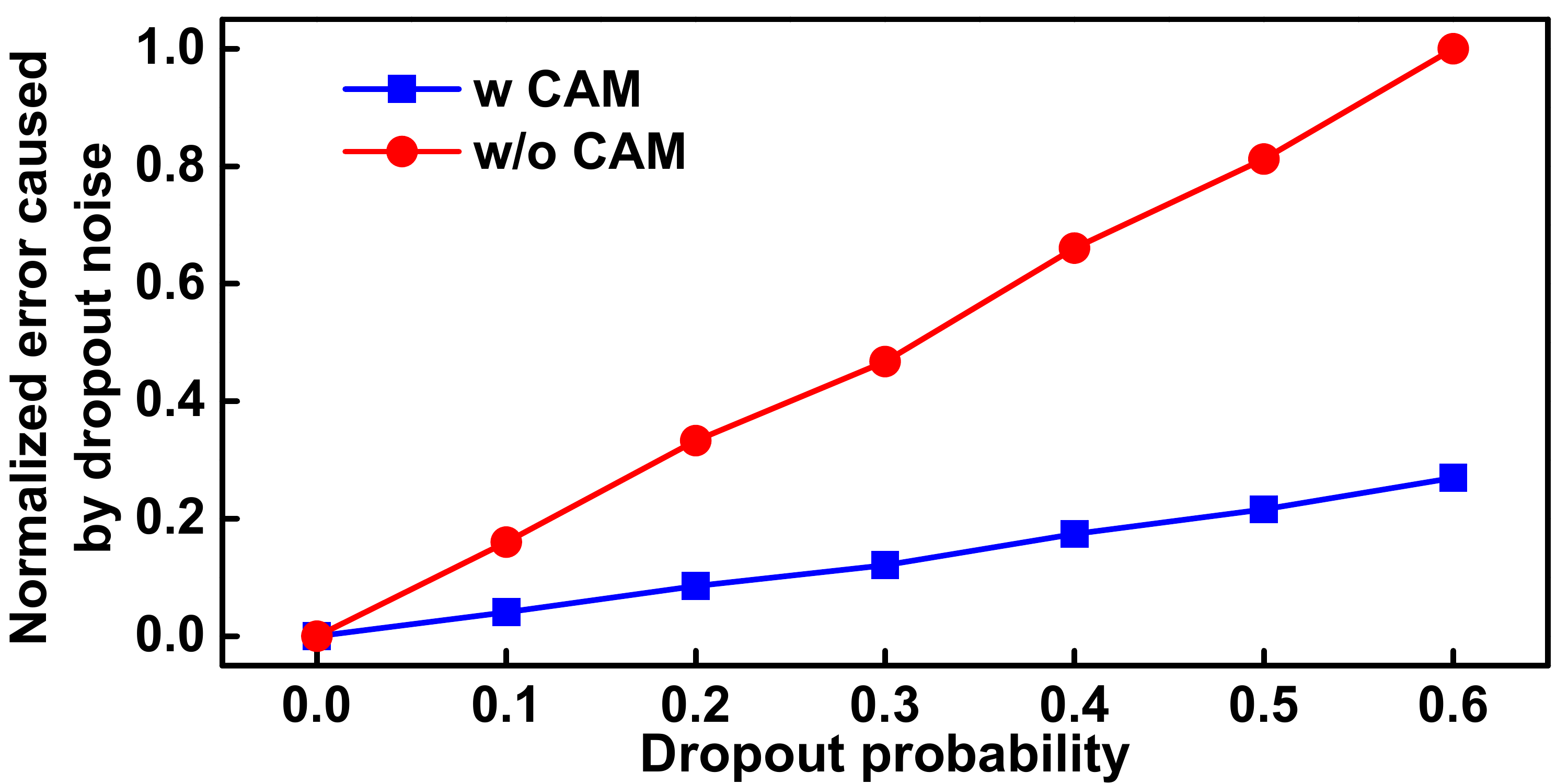}
\caption{We feed a random tensor to a convolutional filter, one with CAM before a $3\times3$ convolution filter and the other one without CAM. We randomly add dropout noise to the input, and measure the output errors. As we increase the dropout probability, the error also increases. For all dropout probabilities, adding CAM improve the robustness towards the dropout noise and therefore, the error is always smaller.}
\vspace{-0.5cm}
\label{fig:CAM_Comparison}
\end{center}
\end{figure}

LiDAR point cloud data contains many missing points, which we refer to as dropout noise, as shown in Fig.~\ref{fig:DomainShift}(b). Dropout noise is mainly caused by 1) limited sensor range, 2) mirror reflection (instead of diffusion reflection) of sensing lasers on smooth surfaces, and 3) jitter of the incident angle. Dropout noise has a significant impact on SqueezeSeg, especially in early layers of a network. At early layers where the receptive field of the convolution filter is very small, missing points in a small neighborhood can corrupt the output of the filter significantly. To illustrate this, we conduct a simple numerical experiment, where we randomly sample an input tensor and feed it into a $3\times3$ convolution filter. We randomly drop out some pixels from the input tensor, and as shown in Fig.~\ref{fig:CAM_Comparison}, as we increase the dropout probability, the difference between the errors of the corrupted output and the original output also increases.

\begin{figure}[!t]
\begin{center}
\centering \includegraphics[width=1.0\linewidth]{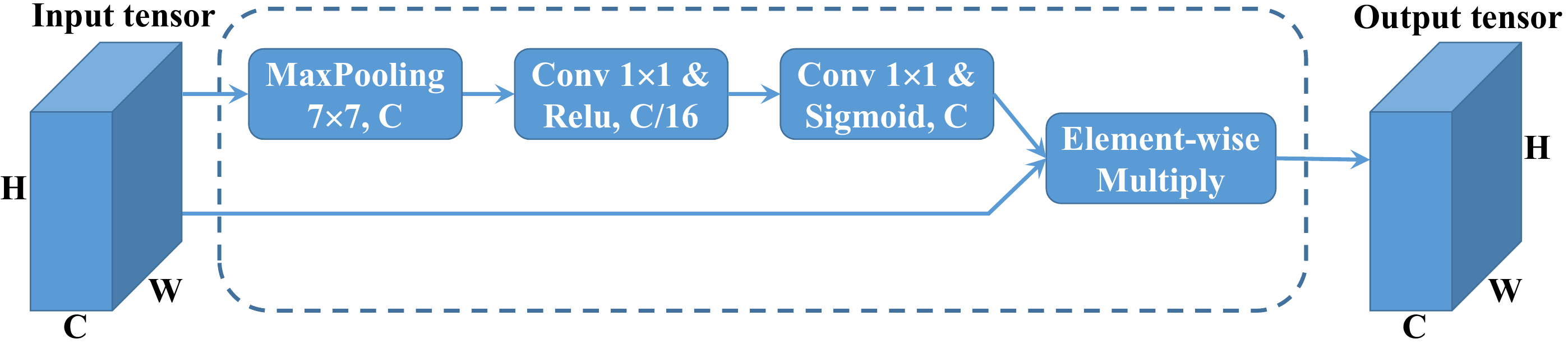}

\caption{Structure of Context Aggregation Module.}
\vspace{-0.6cm}
\label{fig:CAM}
\end{center}
\end{figure}

This problem not only impacts SqueezeSeg when trained on real data, but also leads to a serious domain gap between synthetic data and real data, since simulating realistic dropout noise from the same distribution is very difficult.

To solve this problem, we propose a novel Context Aggregation Module (CAM) to reduce the sensitivity to dropout noise. As shown in Fig.~\ref{fig:CAM}, CAM starts with a max pooling with a relatively large kernel size. The max pooling aggregates contextual information around a pixel with a much larger receptive field, and it is less sensitive to missing data within its receptive field. Also, max pooling can be computed efficiently even with a large kernel size. The max pooling layer is then followed by two cascaded convolution layers with a ReLU activation in between. Following~\cite{hu2018squeeze}, we use the \textit{sigmoid} function to normalize the output of the module and use an element-wise multiplication to combine the output with the input. As shown in Fig.~\ref{fig:CAM_Comparison}, the proposed module is much less sensitive to dropout noise -- with the same corrupted input data, the error is significantly reduced.

In SqueezeSegV2, we insert CAM after the output of the first three modules (1 convolution layer and 2 FireModules), where the receptive fields of the filters are small. As can be seen in later experiments, CAM 1) significantly improves the accuracy when trained on real data, and 2) significantly reduces the domain gap while trained on synthetic data and testing on real data.

\subsection{Focal Loss}
\label{ssec:Focal} 
LiDAR point clouds have a very imbalanced distribution of point categories -- there are many more background points than there are foreground objects such as cars, pedestrians, etc. This imbalanced distribution makes the model focus more on easy-to-classify background points which contribute no useful learning signals, with the foreground objects not being adequately addressed during training.

To address this problem, we replace the original cross entropy loss from SqueezeSeg~\cite{wu2017squeezeseg} with a focal loss~\cite{lin2018focal}. The focal loss modulates the loss contribution from different pixels and focuses on hard examples. For a given pixel label $t$, and the predicted probability of $p_t$, focal loss~\cite{lin2018focal} adds a modulating factor $(1-p_t)^{\gamma}$ to the cross entropy loss. The focal loss for that pixel is thus
\begin{equation}
FL(p_t) =-(1-p_t)^{\gamma}\log{(p_t)}
\end{equation}
When a pixel is mis-classified and $p_t$ is small, the
modulating factor is near 1 and the loss is unaffected. As
$p_t \rightarrow1$, the factor goes to 0, and the loss for well-classified
pixels is down-weighted. The focusing parameter $\gamma$ smoothly adjusts the rate at which well-classified examples are down-weighted.
When $\gamma = 0$, the Focal Loss is equivalent to the Cross Entropy Loss. As $\gamma$ increases, the effect of the modulating factor is likewise increased. We choose $\gamma$ to be $2$ in our experiments.

\subsection{Other Improvements}
\textbf{LiDAR Mask}: Besides the original (x, y, z, intensity, depth) channels, we add one more channel -- a binary mask indicating if each pixel is missing or existing. As we can see from Table~\ref{tab:SqueezeSegV2}, the addition of the mask channel significantly improves segmentation accuracy for cyclists.

\textbf{Batch Normalization}: Unlike SqueezeSeg~\cite{wu2017squeezeseg}, we also add batch normalization (BN)~\cite{ioffe2015batch} after every convolution layer. The BN layer is designed to alleviate the issue of internal covariate shift -- a common problem for training a deep neural network. We observe an improvement in car segmentation after using BN layers in Table~\ref{tab:SqueezeSegV2}.

\section{Domain Adaptation Training}
\label{sec:Stronger}
\begin{figure}[!t]
\begin{center}
\subfigure[Pre-training: Learned Intensity Rendering]{
\label{fig:LIR}
\includegraphics[width=1.\linewidth]{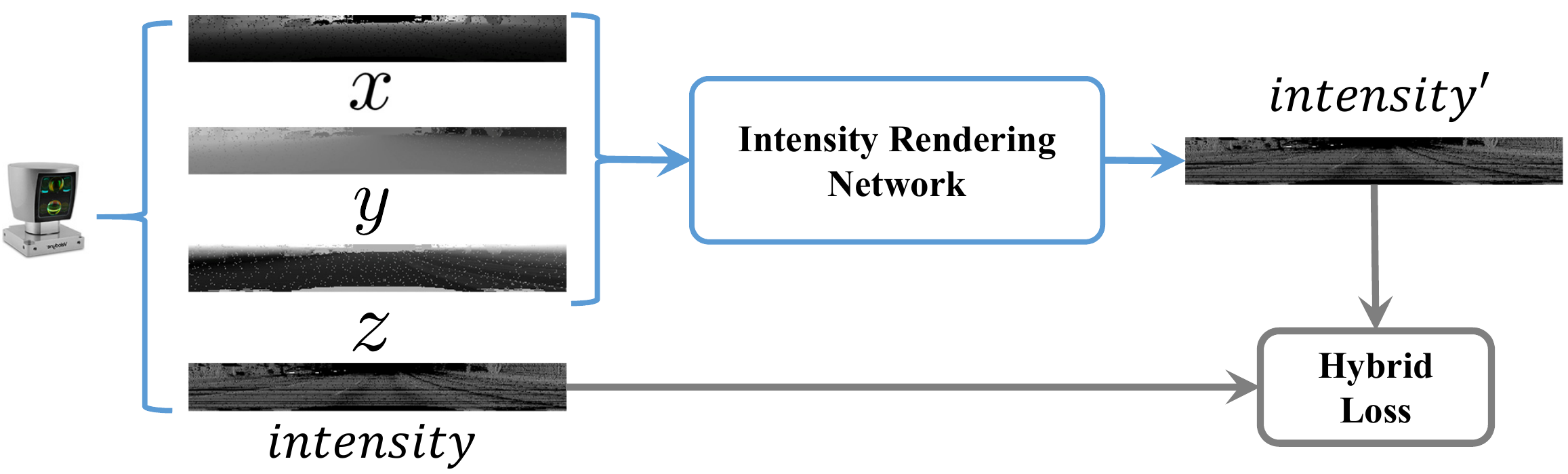}
}
\subfigure[Training: Geodesic Correlation Alignment]{
\label{fig:GCA}
\includegraphics[width=1.\linewidth]{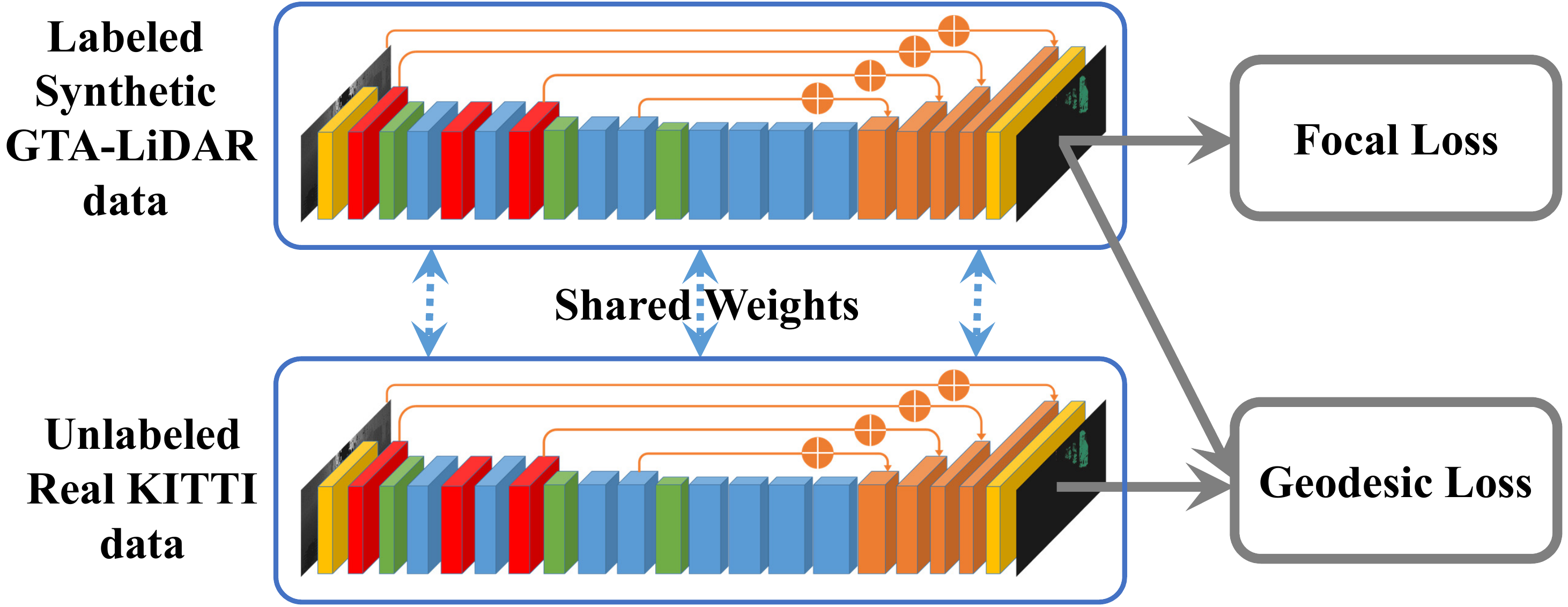}
}
\subfigure[Post-training: Progressive Domain Calibration]{
\label{fig:PDC}
\includegraphics[width=1.\linewidth]{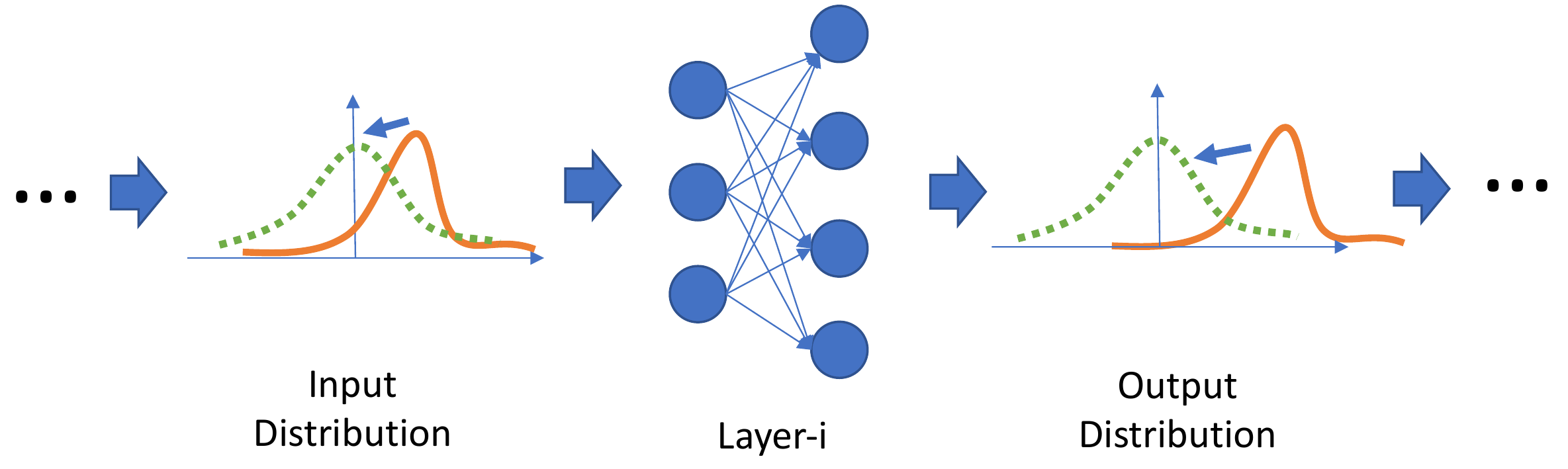}
}
\caption{Framework of the proposed unsupervised domain adaptation method for road-object segmentation from the synthetic GTA-LiDAR dataset to the real-world KITTI dataset.}

\vspace{-0.6cm}
\label{fig:DAFramework}
\end{center}
\end{figure}

In this section, we introduce our unsupervised domain adaptation pipeline that trains SqueezeSegV2 on synthetic data and improves its performance on real data. We construct a large-scale 3D LiDAR point cloud dataset, GTA-LiDAR, with 100,000 LiDAR scans simulated on GTA-V. To deal with the \emph{domain shift} problem, we employ three strategies: learned intensity rendering, geodesic correlation alignment, and progressive domain calibration, as shown in Fig.~\ref{fig:DAFramework}.
\vspace{-0.1cm}

\subsection{The GTA-LiDAR Dataset}
\label{ssec:GTA-LiDAR}
We synthesize 100,000 LiDAR point clouds in GTA-V to train SqueezeSegV2. We use the framework in \cite{philip2018free} to generate depth semantic segmentation maps, and use the method in \cite{yue2018lidar} to do Image-LiDAR registration in GTA-V. Following \cite{yue2018lidar}, we collect 100,000 point cloud scans by deploying a virtual car to drive autonomously in the virtual world. GTA-V provides a wide variety of scenes, car types, traffic conditions, etc., which ensures the diversity of our synthetic data. Each point in the synthetic point cloud contains one label, one distance and $x, y, z$ coordinates. However, it does not contain intensity, which represents the magnitude of the reflected laser signal. Also, the synthetic data does not contain dropout noise as in the real data. Because of such distribution discrepancies, the model trained on synthetic data fails to transfer to real data.

\subsection{Learned Intensity Rendering}
\label{ssec:Intensity}
The synthetic data only contains $x, y, z, depth$ channels and does not have intensity. As shown in SqueezeSeg~\cite{wu2017squeezeseg}, intensity is an important signal. The absence of intensity can lead to serious accuracy loss. Rendering realistic intensity is a non-trivial task, since a multitude of factors that affect intensity, such as surface materials and LiDAR sensitivity, are generally unknown to us.

\begin{figure*}[!t]
\begin{center}
\centering \includegraphics[width=1.0\linewidth]{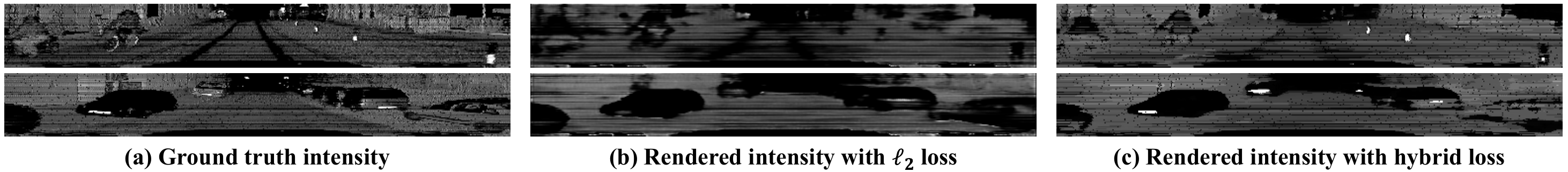}
\caption{Rendered v.s. ground truth intensity in the KITTI dataset.}
\vspace{-10pt}
\label{fig:PredictedIntensity}
\end{center}
\end{figure*}

To solve this problem, we propose a method called \textit{learned intensity rendering}. The idea is to use a network to take the $x, y, z, depth$ channels of the point cloud as input, and predict the intensity. Such rendering network can be trained with unlabeled LiDAR data, which can be easily collected as long as a LiDAR sensor is available. As shown in Fig.~\ref{fig:LIR}, we train the rendering network in a self-supervision fashion, splitting the $x, y, z$ channels as input to the network and the intensity channel as the label. The structure of the rendering network is almost the same as SqueezeSeg, except that the CRF layer is removed. 

The intensity rendering can be seen as a regression problem, where the $\ell_2$ loss is a natural choice. However,  $\ell_2$ fails to capture the multi-modal distribution of the intensity -- given the same input of $x, y, z$, the intensity can differ. To model this property, we designed a hybrid loss function that involves both classification and regression. We divide the intensity into $n=10$ regions, with each region having a reference intensity value. The network first predicts which region the intensity belongs to. Once the region is selected, the network further predicts a deviation from the reference intensity. This way, the categorical prediction can capture the multi-modal distribution of the intensity, and the deviation prediction leads to more accurate estimations. We train the rendering network on the KITTI \cite{geiger2012we} dataset with the hybrid loss function and measure its accuracy with mean squared error (MSE). Compared to $\ell_2$ loss, the converged MSE drops significantly by 3X from 0.033 to 0.011. A few rendered results using two different losses are shown in Fig.~\ref{fig:PredictedIntensity}. After training the rendering network, we feed synthetic GTA-LiDAR data into the network to render point-wise intensities. 

\subsection{Geodesic Correlation Alignment}
\label{ssec:LE}
After rendering intensity, we train SqueezeSegV2 on the synthetic data with focal loss. However, due to distribution discrepancies between synthetic data and real data, the trained model usually fails to generalize to real data.

To reduce this domain discrepancy, we adopt geodesic correlation alignment during training. As shown in Fig.~\ref{fig:GCA}, at every step of training, we feed in one batch of synthetic data and one batch of real data to the network. We compute the focal loss on the synthetic batch, where labels are available. Meanwhile, we compute the geodesic distance \cite{morerio2018minimal} between the output distributions of two batches. The total loss now contains both the focal loss and the geodesic loss. Where the focal loss focuses on training the network to learn semantics from the point cloud, the geodesic loss penalizes discrepancies between batch statistics from two domains. Note that other distances such as the Euclidean distance can also be used to align the domain statistics. However, we choose the geodesic distance over the Euclidean distance since it takes into account the manifold’s curvature. More details can be found in \cite{morerio2018minimal}.

We denote the input synthetic data as $X_{sim}$, synthetic labels as $Y_{sim}$ the input real data as $X_{real}$. Our loss function can be computed as
\begin{equation}
 FL(X_{sim}, Y_{sim})  + \lambda \cdot GL(X_{sim}, X_{real}), 
\end{equation}
where $FL$ denotes focal loss between the synthetic label and network prediction, $GL$ denotes the geodesic loss between batch statistics of synthetic and real data. $\lambda$ is a weight coefficient and we set it to 10 in our experiment. Note that in this step, we only require unlabeled real data, which is much easier to obtain than annotated data as long as a LiDAR sensor is available.

\begin{algorithm}[!t]
{
\small
\KwIn{Unlabeled real data $\mathcal{X}$, model $\mathcal{M}$}
$\mathcal{X}^{(0)} \leftarrow  \mathcal{X}$ \\
\For{layer $l$ in the model $\mathcal{M}$}{
   $\mathcal{X}^{(l)} \leftarrow \mathcal{M}^{(l)}(\mathcal{X}^{(l-1)})$\\
   $\mu^{(l)} \leftarrow \mathbb{E} (\mathcal{X}^{(l)}$), $\sigma^{(l)} \leftarrow \sqrt{Var(\mathcal{X}^{(l)})}$  \\
   Update the BatchNorm parameters of $\mathcal{M}^{(l)}$ \\
   $\mathcal{X}^{(l)} \leftarrow (\mathcal{X}^{(l)} - \mu^{(l)}) / \sigma^{(l)}$  \\
}
\KwOut{Calibrated model $\mathcal{M}$}

}
\small\caption{Progressive Domain Calibration}
\label{Alg:MTSSRLearning}
\end{algorithm}

\subsection{Progressive Domain Calibration}
\label{ssec:PBA}
After training SqueezeSegV2 on synthetic data with geodesic correlation alignment, each layer of the network learns to recognize patterns from its input and extract higher level features. However, due to the non-linear nature of the network, each layer can only work well if its input is constrained within a certain range. Taking the ReLU function as an example, if somehow its input distribution shifts below 0, the output of the ReLU becomes all zero. Otherwise, if the input shifts towards larger than 0, the ReLU becomes a linear function. For deep learning models with multiple layers, distribution discrepancies from the input data can lead to distribution shift at the output of each layer, which is accumulated or even amplified across the network and eventually leads to a serious degradation of performance, as illustrated in Fig. \ref{fig:PDC}.  

To address this problem, we employ a post training procedure called progressive domain calibration (PDC). The idea is to break the propagation of the distribution shift through each layer with progressive layer-wise calibration. For a network trained on synthetic data, we feed the real data into the network. Staring from the first layer, we compute its output statistics (mean and variance) under the given input, and then re-normalize the output's mean to be 0 and its standard deviation to be 1, as shown in Fig.~\ref{fig:PDC}. Meanwhile, we update the batch normalization parameters (mean and variance) of the layer with the new statistics. We progressively repeat this process for all layers of the network until the last layer. Similar to geodesic correlation alignment, this process only requires unlabeled real data, which is presumably abundant. This algorithm is summarized in Algorithm~\ref{Alg:MTSSRLearning}.
A similar idea was proposed in \cite{li2018adaptive}, but PDC is different since it performs calibration progressively, making sure that the calibrations of earlier layers do not impact those of later layers.

\section{Experiments}
\label{sec:Experiments}
In this section, we introduce the details of our experiments. 
We train and test SqueezeSegV2 on a converted KITTI \cite{geiger2012we} dataset as \cite{wu2017squeezeseg}. To verify the generalization ability, we further train SqueezeSegV2 on the synthetic GTA-LiDAR dataset and test it on the real world KITTI dataset.

\begin{figure*}[!t]
\begin{center}
\centering \includegraphics[width=0.9\linewidth]{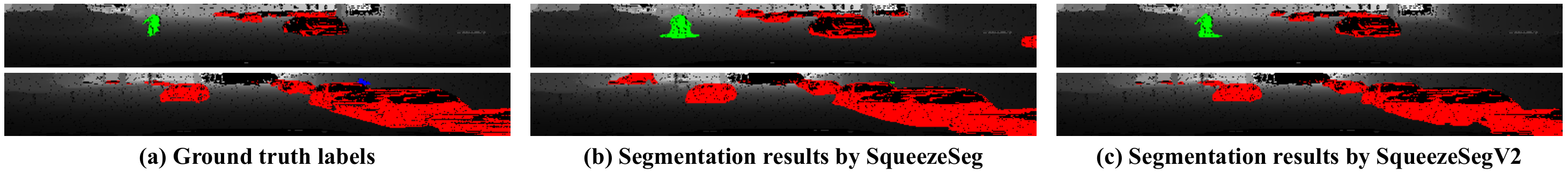}
\vspace{-4pt}
\caption{Segmentation result comparison between SqueezeSeg~\cite{wu2017squeezeseg} and our SqueezeSegV2 (red: car, green: cyclist). Note that in first row, SqueezeSegV2 produces much more accurate segmentation for the cyclist. In the second row, SqueezeSegV2 avoids a falsely detected car that is far away.}
\label{fig:SqueezeSegV1V2}
\vspace{-5pt}
\end{center}
\end{figure*}

\begin{figure*}[!t]
\begin{center}
\centering \includegraphics[width=0.9\linewidth]{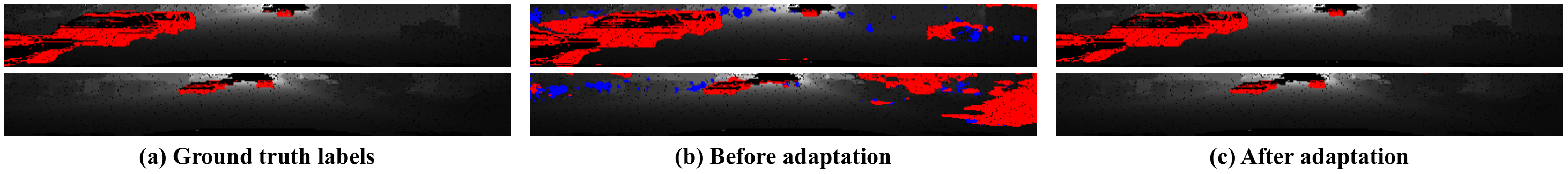}
\vspace{-6pt}
\caption{Segmentation result comparison before and after domain adaptation (red: car, blue: pedestrian).}
\label{fig:DomainAdaptationResult}
\vspace{-12pt}
\end{center}
\end{figure*}

\subsection{Experimental Settings}
\label{ssec:Settings}
We compare the proposed method with SqueezeSeg~\cite{wu2017squeezeseg}, one state-of-the-art model for semantic segmentation from 3D LiDAR point clouds. We use KITTI~\cite{geiger2012we} as the real world dataset. KITTI provides images, LiDAR scans, and 3D bounding boxes organized in sequences. Following~\cite{wu2017squeezeseg}, we obtain the point-wise labels from 3D bounding boxes, all points of which are considered part of the target object. In total, 10,848 samples with point-wise labels are collected. For SqueezeSegV2, the dataset is split into a training set with 8,057 samples and a testing set with 2,791 samples. 
For domain adaptation, we train the model on GTA-LiDAR, and test it on KITTI for comparison.

Similar to~\cite{wu2017squeezeseg}, we evaluate our model's performance on class-level segmentation tasks by a point-wise comparison of the predicted results with ground-truth labels. We employ intersection-over-union (IoU) as our evaluation metric, which is defined as $IoU_c = \frac{|\mathcal{P}_c \cap \mathcal{G}_c|}{|\mathcal{P}_c \cup \mathcal{G}_c|}$, where $\mathcal{P}_c$ and $\mathcal{G}_c$ respectively denote the predicted and ground-truth point sets that belong to class-$c$. $|\cdot|$ denotes the cardinality of a set. 

\begin{table}[!t]
\begin{center}

\caption{Segmentation performance (IoU, \%) comparison between the proposed SqueezeSegV2 (+BN+M+FL+CAM) model and state-of-the-art baselines on the KITTI dataset. }
\begin{tabular}
{c | c c c c}
\hline
& Car & Pedestrian & Cyclist & Average\\
\hline
SqueezeSeg~\cite{wu2017squeezeseg} & 64.6 & 21.8  &  25.1  & 37.2 \\
+BN & 71.6  & 15.2  & 25.4 & 37.4 \\
+BN+M  & 70.0  & 17.1 & 32.3 & 39.8\\
+BN+M+FL   & 71.2 & 22.8  &  27.5 & 40.5 \\
+BN+M+FL+CAM & \textbf{73.2} & \textbf{27.8} & \textbf{33.6} & \textbf{44.9} \\
\hline
PointSeg \cite{wang2018pointseg} & 67.4 & 19.2 & 32.7 & 39.8 \\
\hline
\end{tabular}
\label{tab:SqueezeSegV2}
\end{center}
\textbf{+BN} denotes using batch normalization. \textbf{+M} denotes adding LiDAR mask as input. \textbf{+FL} denotes using focal loss. \textbf{+CAM} denotes using the CAM module.
\end{table}

\subsection{Improved Model Structure}
\label{ssec:Results_SqueezeSegV2}
The performance comparisons, measured in IoU, between the proposed SqueezeSegV2 model and baselines  are shown in Table~\ref{tab:SqueezeSegV2}. Some segmentation results are shown in Fig.~\ref{fig:SqueezeSegV1V2}.

From the results, we have the following observations. (1) both batch normalization and the mask channel can produce better segmentation results - batch normalization boosts segmentation of cars, whereas the mask channel boosts segmentation of cyclists. (2) Focal loss improves segmentation of pedestrians and cyclists. The number of points corresponding to pedestrians and cyclists is low relative to the large number of background points. This class imbalance causes the network to focus less on the pedestrian and cyclist classes. Focal loss mitigates this problem by focusing the network on optimization of these two categories. (3) CAM significantly improves the performance of all the classes by reducing the network's sensitivity to dropout noise. 

\subsection{Domain Adaptation Pipeline}
\label{ssec:Results_SqueezeSegUDA}

\begin{table}[!t]
\begin{center}
\small
\caption{Segmentation performance (IoU, \%) of the proposed domain adaptation pipeline from GTA-LiDAR to the KITTI.}
\begin{tabular}
{c | c  c }
\hline
& \multicolumn{1}{c}{Car} & \multicolumn{1}{c}{Pedestrian}\\
\hline
SQSG trained on GTA \cite{wu2017squeezeseg}  &  29.0 & -  \\
SQSG trained on GTA-LiDAR & 30.0 & 2.1 \\
+LIR   &  42.0 & 16.7  \\
+LIR+GCA &  48.2 & 18.2\\
+LIR+GCA+PDC  &  50.3 & 18.6  \\
+LIR+GCA+PDC+CAM  &  \textbf{57.4}  & \textbf{23.5}  \\
\hline
SQSG trained on KITTI w/o intensity \cite{wu2017squeezeseg} & 57.1 & - \\
\hline
\end{tabular}
\label{tab:DA_Result}
\end{center}
 \textbf{SQSG} denotes SqueezeSeg. \textbf{+LIR} denotes using learned intensity rendering. \textbf{+GCA} denotes using geodesic correlation alignment. \textbf{+PDC} denotes using progressive domain calibration. \textbf{+CAM} denotes using the CAM module.
 
 \vspace{-0.4cm}
\end{table}

The performance comparisons, measured in IoU, between the proposed domain adaptation pipeline and baselines  are shown in Table~\ref{tab:DA_Result}. Some segmentation results are shown in Fig.~\ref{fig:DomainAdaptationResult}. From the results, we have the following observations. (1) Models trained on the source domain without any adaptation does not perform well. Due to the influence of \emph{domain discrepancy}, the joint probability distributions of observed LiDAR and road-objects greatly differ in the two domains. This results in the model's low transferability from the source domain to the target domain. (2) All adaptation methods are effective, with the combined pipeline performing the best, demonstrating its effectiveness. (3) Adding the CAM to the network also significantly boosts the performance on the real data, supporting our hypothesis that dropout noise is a significant source of domain discrepancy. Therefore, improving the network to make it more robust to dropout noise can help reduce the domain gap. (4) Compared with \cite{wu2017squeezeseg} where a SqueezeSeg model is trained on the real KITTI dataset but without intensity, our SqueezeSegV2 model trained purely on synthetic data and unlabeled real data achieves a better accuracy, showing the effectiveness of our domain adaptation training pipeline. (5) Compared with our latest SqueezeSegV2 model trained on the real KITTI dataset, there is still an obvious performance gap. Adapting the segmentation model from synthetic LiDAR point clouds is still a challenging problem.

\section{CONCLUSION}
\label{sec:Conclusion}
In this paper, we proposed SqueezeSegV2 with better segmentation performance than the original SqueezeSeg and a domain adaptation pipeline with stronger transferability. We designed a context aggregation module to mitigate the impact of dropout noise. Together with other improvements such as focal loss, batch normalization and a LiDAR mask channel, SqueezeSegV2 sees accuracy improvements of 6.0\% to 8.6\% in various pixel categories over the original SqueezeSeg. We also proposed a domain adaptation pipeline with three components: learned intensity rendering, geodesic correlation alignment, and progressive domain calibration. The proposed pipeline significantly improved the real world accuracy of the model trained on synthetic data by 28.4\%, even out-performing a baseline model \cite{wu2017squeezeseg} trained on the real dataset.

\section*{Acknowledgement}
This work is partially supported by Berkeley Deep Drive (BDD), and partially sponsored by individual gifts from Intel and Samsung. We would like to thank Alvin Wan and Ravi Krishna for their constructive feedback.

\addtolength{\textheight}{-8cm}   


\newpage
\small\bibliographystyle{IEEEtran}
\bibliography{root}

\end{document}